\documentclass[sigconf]{acmart}
\def\BibTeX{{\rm B\kern-.05em{\sc i\kern-.025em b}\kern-.08emT\kern-.1667em\lower.7ex\hbox{E}\kern-.125emX}}

\usepackage[utf8]{inputenc} 
\usepackage[T1]{fontenc}    
\usepackage{hyperref}       
\usepackage{url}            
\usepackage{booktabs}       
\usepackage{amsfonts}       
\usepackage{nicefrac}       
\usepackage{microtype}      

\usepackage{bm,amsmath}
\usepackage{widetext}
\usepackage{multicol,lipsum,multirow}

\usepackage{xspace}
\usepackage{color}
\usepackage{natbib}
\usepackage[toc,page]{appendix}
\usepackage{subcaption}
\usepackage{sidecap}
\sidecaptionvpos{figure}{c}
\usepackage{wrapfig}
\usepackage{booktabs} 
\usepackage{graphicx} 
\usepackage{csquotes,xspace,algorithmic,enumitem,color}
\usepackage[linesnumbered,ruled]{algorithm2e}

\newcommand{\Hb}{\mathbf{H}}

\newcommand{\eb}{\mathbf{e}}

\newcommand{\hb}{\mathbf{h}}

\newcommand{\xseq}{x_{n,1:\text{T}_n}}
\newcommand{\eye}{\mathbf{I}}
\newcommand{\real}[1]{\mathbb{R}^{#1}}
\newcommand{\data}{\mathcal{D}}
\newcommand{\normal}{\mathcal{N}}
\newcommand{\E}{\mathbb{E}}
\newcommand{\hide}[1]{}
\newcommand{\KL}[2]{\text{KL}(#1 \text{ } || \text{ }#2)}
\newcommand{\ent}[1]{\mathbb{H}({#1})}
\newcommand{\elbo}{\mathcal{L}}
\newcommand{\squash}{\xi}
\newcommand{\grad}[1]{\nabla_{#1}}
\newcommand{\es}{s}

\newcommand{\mname}{\texttt{EVA}\xspace}
\newcommand{\cmname}{\texttt{EVA}$_c$\xspace}

\acmDOI{10.475/123_4}

\begin{document}

\title{\mname: Generating Longitudinal Electronic Health Records Using Conditional Variational Autoencoders}

\author{Siddharth Biswal}
\affiliation{%
  \institution{Georgia Institute of Technology}
}

\author{Soumya Ghosh}
\affiliation{%
  \institution{IBM Research}
}

\author{Jon Duke}
\affiliation{%
  \institution{Georgia Institute of Technology}
}

\author{Bradley Malin}
\affiliation{%
 \institution{Vanderbilt University	}
}

\author{Walter Stewart}
\affiliation{%
  \institution{Sutter Health}
  }

\author{Jimeng Sun}
\affiliation{%
  \institution{University of Illinois Urbana-Champaign}
}

\begin{abstract}
Researchers require timely access to real-world longitudinal electronic health records (EHR) to develop, test, validate, and implement machine learning solutions that improve the quality and efficiency of healthcare. In contrast, health systems value deeply patient privacy and data security. De-identified EHRs do not adequately address the needs of health systems, as de-identified data are susceptible to re-identification and its volume is also limited. Synthetic EHRs offer a potential solution. In this paper, we propose EHR Variational Autoencoder (\mname) for synthesizing sequences of discrete EHR encounters (e.g., clinical visits) and encounter features (e.g., diagnoses, medications, procedures). We illustrate that \mname can produce realistic EHR sequences, account for individual differences among patients, and can be conditioned on specific disease conditions, thus enabling disease-specific studies. We design efficient, accurate inference algorithms by combining stochastic gradient Markov Chain Monte Carlo with amortized variational inference. We assess the utility of the methods on large real-world EHR repositories containing over $250,000$ patients. Our experiments, which include user studies with knowledgeable clinicians, indicate the generated EHR sequences are realistic. We confirmed the performance of predictive models trained on the synthetic data are similar with those trained on real EHRs. Additionally, our findings indicate that augmenting real data with synthetic EHRs results in the best predictive performance - improving the best baseline by as much as $8\%$ in top-20 recall.

\end{abstract}

\maketitle

\section{Introduction}
Electronic health records (EHR) are now widely adopted in the US by more than $90\%$ of hospitals and $72\%$ of ambulatory practices ~\cite{henry2016adoption}, opening opportunities to advance development of machine learning (ML) and digital health solutions that improve the quality and efficiency of care. Access to EHR data is essential to the development, design, testing, validation, and implementation of ML solutions. But, health systems are often reluctant to share EHR data for research and development as they ultimately assume responsibility for the risk and the related consequences of data breaches. More often than not, health systems simply refuse to provide access to data or, alternatively, impose significant administrative burdens that can be costly (e.g., indemnity insurance) or result in serious delays that threaten the success of research endeavors. 

While healthcare is on the precipice of profound changes, mediated by artificial intelligence, automation, and other means of transforming care ~\cite{musen2014clinical, choi2016doctor, futoma2017improved, JHO14, YWang15}, a fundamental conflict stands in the way between health systems that control access to EHR data and researchers who need the data. A health system’s need for certainty of data security is in direct conflict with a researcher’s need for direct access to data that accurately represents the longitudinal history of large patient population. Current methods for resolving the conflict favor the needs of one or the other parties, not the joint interests of both. The standard method for alleviating EHR data privacy rely on methods such as de-identification.  Moreover, even carefully de-identified EHRs are still susceptible to re-identification attacks ~\cite{el2011systematic, el2015anonymising}. And the volume of the de-identified EHRs are bounded by the original EHR data. Especially, if one wants to study patients with specific condition combinations, the available EHR data can be very limited. 

Ideally, synthetic EHRs can offer a potential solution as they yield a database that is beyond de-identification hence immune to re-identification, while preserving temporal patterns in real longitudinal EHRs. More generally, synthetic EHR methods hold the promise to create large, realistic EHR datasets that address the needs of researchers and ensure complete security for health systems. 
Recent efforts that leverage deep generative models for synthesizing EHRs~\citep{choi2017generating, beaulieu2017privacy, baowaly2018synthesizing},  while promising, are limited by their inability to generate sequences, instead only generating a static patient representation without temporal variation.
While significant progress has been made in generating continuous data such as images ~\cite{goodfellow2014generative} and audio ~\cite{van2016wavenet}, generation of realistic discrete sequences, even natural language text fragments, remains an open problem. In order to generate realistic longitudinal EHRs, statistical models must account for such patient-level differences. Finally, it is often of interest to generate EHRs of a cohort of patients suffering from either a single or a collection of pre-specified medical conditions.

In this paper, we develop EHR Variational Autoencoder (\mname), a deep generative model to address these challenges. Focusing on the problem of generating realistic discrete EHR code sequences, we advance the field in several ways, 
\begin{itemize}[leftmargin=*]
\item {\bf Temporal conditional generation:} We develop deep generative models that provide conditional generation of EHR sequences specific to medical conditions of interest. 
\item {\bf Diverse sequence generation: } We  retain uncertainty in the parameters of the model leads to diversity in sequences and is crucial for generating realistic synthetic EHRs. 
\item {\bf Efficient generation algorithm:} We design a new inference algorithm that combines stochastic gradient Markov chain Monte-Carlo (SGMCMC) and amortized variational inference. This allows us to efficiently perform full posterior inference over the parameters of the model without sacrificing predictive accuracy. 
\item {\bf Comprehensive evaluation:} Through large scale experiments on real EHR repositories containing more than $250,000$ patients and over \emph{13 million} visits, we thoroughly vet the proposed methods and demonstrate their efficacy in generating realistic EHR sequences. We conduct user study with a physician and also demonstrate that the generated EHRs are realistic through predictive modeling tasks. We find that predictive models trained on datasets augmented with synthetic EHRs improve upon those trained without, by as much as $8\%$ in top-20 recall. 
\end{itemize}

\section{Generative Models for Electronic Health Records}\label{sec:back}
To formalize the system, we assume the data is comprised of a cohort of $N$ patients. Each patient is represented by a variable length sequence $\xseq = \{x_{n,1}, x_{n,2}, \ldots, x_{n,\text{T}_n} \}$ of $\text{T}_n$ visits to a health care provider. Each visit $x_{n,t}$ is a V-dimensional binary vector, where $x_{n, t}[v] = 1$ if the $v^\text{th}$ code for patient $n$ was observed at visit $t$ and $0$ otherwise. V denotes the cardinality of the set of possible codes. 

We consider directed latent variable models endowed with autoregressive likelihoods~\cite{bowman2015generating} for modeling $\xseq$. Here, each patient is modeled with a single latent variable and the sequential dependencies in her records are captured via autoregressive likelihoods. Since the likelihoods specify an explicit parametric distribution over the observed records, the discreteness of $\xseq$ presents no particular challenge to learning. 
%
%
\paragraph{Variational Autoencoders (VAEs)}\label{sec:vae} refer to a particular combination of a latent variable model and an amortized variational inference scheme~\cite{kingma2013auto}. Consider a statistical model that specifies the marginal distribution of an observed data instance $x$ via a parameterized transformation of a latent variable $z$, $p(x; \theta) = \int p(x \mid z; \theta)p(z) dz$. The parameters $\theta$ are shared amongst all data instances, while the latent variables $z$, typically endowed with standard Gaussian priors $p(z) = \normal(0, \eye)$, are data instance specific. Given a collection of instances, $\data = \{x_1,\ldots, x_n\}$, the model can be learned by maximizing the marginal likelihood $p(\data\mid \theta)$ with respect to $\theta$. Unfortunately, the marginalization over the latent variables is, in general, intractable. To cope with fact, variational inference instead maximizes a tractable lower bound to the marginal likelihood,
\begin{equation}
\begin{aligned}[b]
& p(\data; \theta) \geq \elbo(\theta, \phi) \\
& = \sum_n \E_{q(z_n \mid x_n; \phi)}[\text{ln } p(x_n \mid z_n; \theta)] - \KL{q(z_n \mid x_n; \phi)}{p(z_n)},
\label{eq:vae}
\end{aligned}
\end{equation}
where $q(z_n \mid x_n; \phi)$ is a tractable surrogate to the true posterior $p(z_n \mid x_n)$. Variational autoencoders amortize the cost of inferring $\{z_1, \ldots, z_N\}$ by using an inference network shared across all data instances to parameterize the approximate distribution, $q(z_n \mid x_n; \phi)$, where $\phi$ denotes the parameters of the inference network. The parameters of the model and the inference network, $\theta$ and $\phi$, are learned jointly by maximizing Equation~\ref{eq:vae}. Owing to the architectural similarity with standard autoencoders~\cite{DAckley1987}, the inference network and the generative model are sometimes referred to as the encoder and the decoder, respectively.
\begin{figure*}[t]
\centering
\includegraphics[width=.65\textwidth]{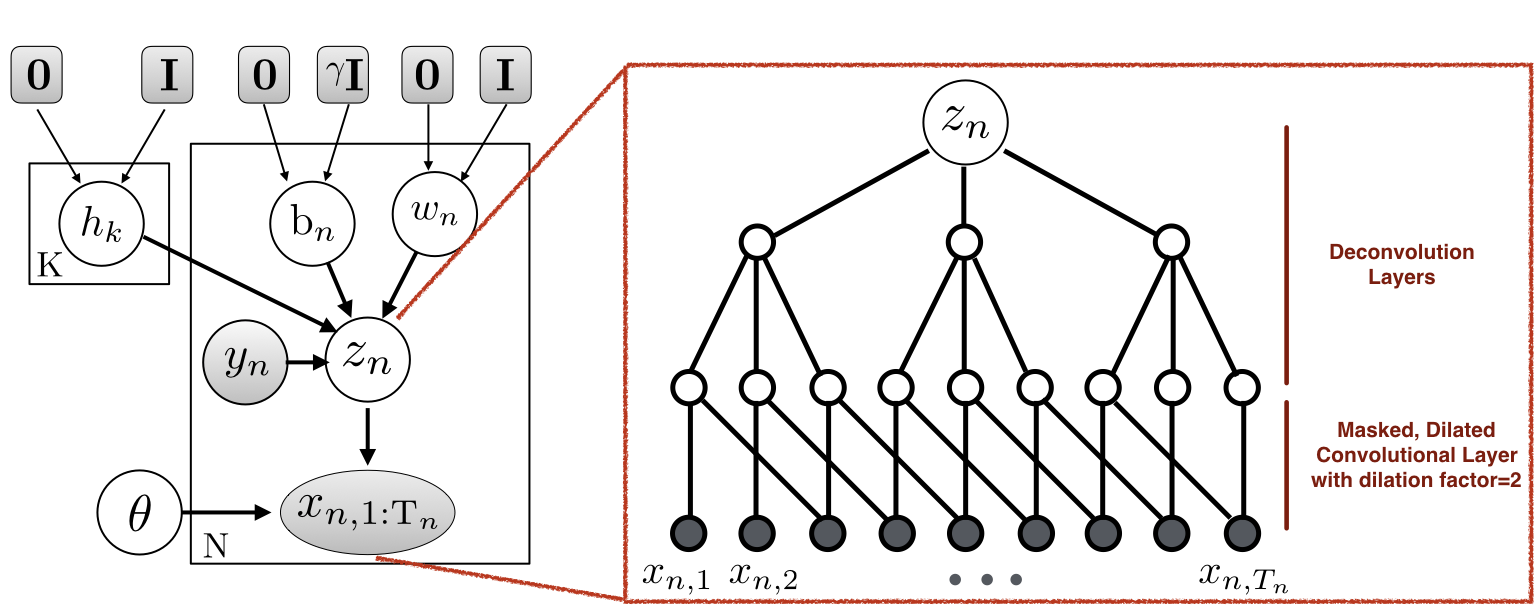}
\caption{\small{\textbf{Graphical model} summarizing the conditional dependencies assumed by the proposed conditional model \cmname. Patient specific latent variables $z_n$ are defined as a noisy linear combination of disease specific representations, $h_k$, $z_n \sim \normal(\Hb\pi_n + b_n, \tau\eye)$. Patients are allowed to deviate from the population through individual biases $b_n$ and by using $\pi_n = y_n \odot \sigma(w_n)$ to weight differently distinct disease representations. $y_n$ is an observed binary vector indicating the diseases afflicting patient $n$ and $\sigma(w_n)$ denotes a logistic transformation applied to each element of $w_n$. The EHR sequence $\xseq$ is generated by transforming $z_n$ using the network architecture shown on the right. The parameters of the network are collectively denoted by $\theta$.} }
\label{fig:evae_gm}
\end{figure*}
%
%

\textbf{Sequential data} VAEs have been extended to modeling sequence collections, $\data = \{x_{1, 1:\text{T}_1}, \ldots, x_{\text{N}, 1:\text{T}_\text{N}} \}$, and  are relevant to modeling  clinical encounters within patients are sequences.  A popular approach~\cite{bowman2015generating} is to retain a sequence-wide latent variable $z_n$ and to parameterize the conditional distribution $p(\xseq \mid z_n; \theta)$ with powerful autoregressive models, such that the distribution over the elements of a sequence is given by $\prod_t p(x_{n, t} \mid x_{n, 1}, \ldots, x_{n, t-1}, z_n; \theta)$. Various flavors of autoregressive architectures for parameterizing the likelihoods have been explored, including long short-term memory networks~\cite{bowman2015generating}, one-dimensional deconvolutional networks~\cite{miao2016language}, and masked and dilated convolution variants~\cite{yang2017improved}, which were originally proposed for sequence-to-sequence machine translation. 
\subsection{Electronic health record Variational Auto-encoders (\mname)}
\label{sec:eva}
Given $N$ longitudinal EHRs, $\mathcal{D} = \{x_{1, 1:\text{T}_1}, \ldots, x_{\text{N}, 1:\text{T}_\text{N}} \}$,
each patient is generated as
\begin{equation}
z_n \sim \normal(\mathbf{0}, \eye); \;\;
\xseq \mid z_n, \theta \sim p(\xseq \mid f_\theta(z_n)),
\end{equation}
where $z_n \in \real{D}$ and $\eye$ is a $D\times D$ identity matrix. The likelihoods are parameterized by a non-linear function $f_\theta$ --- a composition of a series of  one-dimensional deconvolutions and convolutions, parameterized by a set of parameters collectively denoted $\theta$ (Figure~\ref{fig:evae_gm}). The deconvolution operations progressively up-sample the latent representation $z_n$ to match the temporal resolution of the sequence, $\xseq$. The sequential dependencies in $\xseq$ are then modeled via a series of masked, and dilated 1-D convolutions on the up-sampled representation. Dilation allows the effective receptive field size to grow exponentially with depth, while the masking operation ensures that $x_{n, t}$ is independent of future observations $x_{n, >t}$,
\begin{equation}
\begin{aligned}
\small
& p(\xseq \mid f_\theta(z_n)) = p(x_{n, 1} \mid \squash(f_\theta(z_n))) \\
& \prod_{t=2}^{\text{T}_n} p(x_{n, t} \mid \squash(f_\theta(x_{n, t-1}, \ldots, x_{n, t-\es}, z_n))),
\label{eq:lik}
\end{aligned}
\end{equation}
where $\es =  (u - 1) \sum_\ell(d_\ell) + 1$ is the number of past observations we condition on and is specified as a function of the convolutional kernel size $u$, and dilation factor $d_\ell$, for layer $\ell$. In a preprocessing (see the supplement for details and alternatives) step, we group frequently co-occurring codes and model each visit $x_{n, t}$ as a categorically distributed random variable, $ p(x_{n, t} \mid \squash(.)) = \text{Categorical}(x_{n, t} \mid \squash(.))$, where $\squash$ is the softmax function.  

Our choice of $f$ is inspired by recent work that has found similar architectures to be effective at text generation~\cite{semeniuta2017hybrid, yang2017improved}. However, our approach differs in that we endow $\theta$ with its own prior distribution $p(\theta)$. 
We find that this simple modification when combined with an inference algorithm that infers a full posterior distribution over $\theta$ rather than point estimates, leads to improved generative performance. We place a standard normal prior over each element of $\theta$, $\theta_i \sim \normal(0, 1)$. Based on this forumlation, we can summarize the joint distribution (Figure~\ref{fig:evae_gm}) as
\begin{equation}
\begin{aligned}[b]
& p(\mathcal{D}, \{z_n\}_{n=1}^N, \theta) =  p(\theta)\prod_{n=1}^N p(z_n)p(x_{n, 1} \mid \squash(f_\theta(z_n))) \\
& \prod_{t=2}^{\text{T}_n} p(x_{n, t} \mid \squash(f_\theta(x_{n, t-1}, \ldots, x_{n, t-\es}, z_n))).
\end{aligned}
\end{equation}

\subsection{Hierarchically Factorized Conditional (\cmname)}
\label{sec:evac}
\mname as we will see in the experiments, is able to produce realistic EHR sequences. However, it has a few shortcomings. First, \mname does not allow for controlled generation of sequences. While generic EHR sequences are interesting, it is far more useful to have the ability to generate sequences of patients suffering from specific medical conditions, for instance, heart failure or breast cancer due to their significant clinical impact and complex etiology. A second shortcoming stems from the fact that the latent representations $z_n$ are responsible for modeling both differences between different clinical conditions, as well as individual differences between patients with identical clinical conditions. \mname thus lacks knobs for fine grained control of the generation process --- one cannot use EVA to generate patients who all suffer from the same set of ailments but with different severities. 

In this section, we address these shortcomings through a hierarchically factorized conditional variant of \mname, denoted \cmname. We assume that each patient has an available set of meta-data indicating the set of clinical conditions of interest (e.g., whether patient has heart failure or not) in addition to their EHR sequence. We encode this information through a K-dimensional binary vector $y_n$, where $y_{n,k} = 1$ indicates that patient $n$ was diagnosed with condition $k$ and K denotes the number of clinical conditions of interest. 
We represent each condition using a D-dimensional latent variable $h_k \sim \normal(\mathbf{0}, \eye_D)$ and share, $\Hb = [h_1, \ldots, h_{\text{K}}] \in \real{\text{D}\times\text{K}}$, them across patients. Sparse linear combinations of $h_1, \ldots, h_\text{K}$ then engender patient specific representations $z_n$. The sparsity stems from $y_n$ and ensures that only representations that are expressed by patient $n$ contribute to $z_n$. To model the heterogeneity among individuals, we allow patient specific weightings $w_n \sim \normal(\mathbf{0}, \eye_K)$ of the condition specific representations. Furthermore, we allow the patient representations to systematically vary from the mean through patient-specific biases $b_n \sim \normal(\mathbf{0}, \gamma\eye_K)$, where $\gamma$ is a hyper-parameter controlling the scale of the bias
\begin{equation}
z_n = \Hb\pi_n + b_n + \epsilon, \quad \epsilon \sim \normal(0, \tau\eye_D), \quad
\pi_n = y_n \odot \sigma(w_n). 
\end{equation}
Here, $\sigma(w_n)_k = (1 + e^{-w_{nk}})^{-1}$ denotes an element-wise logistic function, and $\odot$ denotes a element-wise multiplication. The non-zero elements of $\pi$, $\pi_{n,k}$ may be interpreted as the intensity with which patient $n$ expresses condition $k$. The constraint of these intensities to the $[0, 1]$ interval both stabilizes training and renders the model more interpretable (by ensuring all intensities are positive). Finally, the conditional distribution $p(\xseq\mid z_n)$ is defined identically to \mname, Equation~\ref{eq:lik}.
The model can then be summarized as (Figure~\ref{fig:evae_gm}) 
\begin{equation}
\begin{aligned}[b]
\small
p(\mathcal{D}, \mathcal{Z}, \theta,& \Hb \mid \eta) = 
p(\theta)p(\Hb)\prod_{n=1}^N p(w_n)p(b_n\mid\gamma)\\
& p(z_n\mid \Hb, \pi_n, b_n, \tau) 
p(\xseq\mid f_\theta(z_n)),
\end{aligned}
\end{equation}
where $\eta = \{\{y_n\}_{n=1}^N, \tau, \gamma$\} and $\mathcal{Z} = \{z_n, w_n, b_n\}_{n=1}^N$. We set both the scale hyperparameters $\tau$ and $\gamma$ to $0.1$. A small $\tau$ encourages the model to explain the observations via the linear combination of condition specific codes, $\Hb\pi_n + b_n$,  rather than white noise. Similarly, a small $\gamma$ value encodes the prior belief that patients only exhibit small systematic differences from the cohort at large. Given a trained model, we use ancestral sampling to generate synthetic sequences.  

We end this section by contrasting hierarchical conditional \mname against supervised VAE~\cite{kingma2014semi}, $z_n \mid p(z)$, $y_n \sim p(y)$, $x_n \mid z_n, y_n \sim p(x_n \mid f(y_n, z_n; \theta))$, a popular conditional variant of the VAE. In the supervised VAE the latent variable $z_n$ must account for both effects arising from different medical conditions as well as those arising from individual differences among patients, learning an entangled representation. Further, $p(y)$ is typically assumed to be a categorical distribution. This makes it difficult to model patients that exhibit more than one condition. In general, such patients can be modeled by the supervised VAE by only resorting to exponentially large representations of $y_n$. In contrast, $\mname_c$ is able to efficiently model such patients while learning population wide medical condition representations in addition to patient-specific representations.

\section{Learning and Inference}

To address the accuracy and efficiency 
goals, we find it useful to treat  the patient-specific latent variables, $z_n, w_n, b_n$, separately from the global variables shared across all patients, $\Hb$ and $\theta$. The former grow with the dataset, and for larger datasets their sheer numbers pose a significant computational challenge. Moreover, we need to infer these latent variables for patients encountered at test time and, thus, require the inference process for these variables to be particularly efficient.

\noindent{\bf Estimating patient-specific variables $z_n, w_n, b_n$:} We adopt amortized variational inference~\cite{gershman2014amortized}. Similar to VAE, we employ inference networks that allow us to amortize the cost of inference across patients. We begin by approximating the posterior over these latent variables with the following tractable approximation,
\begin{equation}
\begin{aligned}[b]
& q(z, w, b \mid \data, \{y_n\}_{n=1}^N)  = \prod_{n=1}^N q(z_n, w_n, b_n \mid \xseq, y_n) \\
& = \prod_{n=1}^N \prod_{a \in \{z, w, b\}}q_{\phi_a}(a_n \mid \xseq, y_n),  
\label{eq:approx_family}
\end{aligned}
\end{equation}
and using inference networks that condition on both the sequence $\xseq$ and the clinical condition vector $y_n$. The conditioning could be implemented by feeding the inference network with a concatenation $\xseq$ and $y_n$. However, such an approach is problematic because of the sequential nature of $\xseq$. It is unclear whether $y_n$ should be concatenated to every $x_{n, t} \in \xseq$ or to some pre-specified $x_{n, t_*} \in \xseq$. We circumvent such issues by instead adopting a product-of-experts parameterization of the variational approximation, $q_{\phi_z}(z_n \mid \xseq, y_n) = q(z_n\mid\xseq)q(z_n\mid y_n) = \normal(z_n \mid \text{lstm}_\mu(\xseq), \text{lstm}_\sigma(\xseq)) \normal(z_n \mid \text{mlp}_\mu(y_n), \text{mlp}_\sigma(y_n))$,
where we employ a bi-directional long short term memory (bi-lstm) network parameterized diagonal Gaussian to represent $q(z_n \mid x_n)$ and a feed-forward multi-layer perceptron (mlp) parameterized diagonal Gaussian to represent $q(z_n \mid y_n)$. Following standard practice, we specify the diagonal variance parameters through a soft-plus transformation of the network outputs. It is then straightforward to combine the two distributions, by noting, $\normal(a\mid \mu, \Sigma) = \normal(a \mid \mu_1, \Sigma_1)\normal(a \mid \mu_2, \Sigma_2)$, where $\Sigma^{-1} = \Sigma_1^{-1} + \Sigma_2^{-1}$ and $
\Sigma^{-1}\mu = \Sigma_1^{-1}\mu_1 + \Sigma_2^{-1}\mu_2$. 
We define the variational approximations for $b_n$ and $w_n$ analogously, arming them with their own bi-lstm and mlp inference networks. Similar product of experts inference networks~\cite{wu2018multimodal} have previously been used for multi-modal learning.

\noindent{\bf Estimating global variables $\Hb$ and $\theta$:} Since the global variables $\Hb$ and $\theta$ are shared across patients, amortization is unnecessary. Instead of limiting ourselves to crude approximations for $\Hb$ and $\theta$, we explore the the full posterior over these variables through stochastic gradient Markov chain Monte Carlo (SG-MCMC). In order to proceed, SG-MCMC methods need gradients of the marginal density, $p(\data, \theta, \Hb \mid \eta)$ with respect to $\Hb$ and $\theta$, 
\begin{equation}
\begin{aligned}[b]
\small
&\grad{\theta, \Hb} \text{ ln }p(\data, \theta, \Hb \mid \eta) \\
& = \grad{\theta, \Hb} \text{ ln } \int p(\data, \theta, \Hb, \{w_n, z_n, b_n\}_{n=1}^N \mid \eta) dw_ndz_ndb_n \nonumber \\
& \underset{\sim}{\propto} \sum_{n=1}^N \grad{\theta} \text{ ln }p(\xseq, \theta \mid z_n^s) \\
& + \grad{\Hb} \text{ ln } p(z_n^s \mid b_n^s, y_n, \Hb) + \grad{\theta} \text{ ln }p(\theta) + \grad{\Hb} \text{ ln }p (\Hb), 
\label{eq:psgld}
\end{aligned}
\end{equation}
where, $w_n^s \sim q_{\phi_w}(w_n\mid \xseq, y_n)$, $b_n^s \sim q_{\phi_b}(b_n\mid \xseq, y_n)$, and $z_n^s \sim q_{\phi_z}(z_n\mid \xseq, y_n)$. Then, we arrive at the approximate proportionality through a single sample importance sampling estimate of the intractable integrals over $w_n$, $b_n$, and $z_n$. The derivation is available in the supplement. With the gradients in hand, we use preconditioned-SGLD~\cite{li2016preconditioned} to sample from the marginal posterior, $p(\theta$, $\Hb\mid \data, \eta)$.
%
%

Finally,
to learn the inference network parameters, we minimize the Kullback–Leibler divergence between the variational approximation specified in Equation~\ref{eq:approx_family} and the marginal posterior over patient-specific variables $\displaystyle\text{KL}(q(w, z, b\mid \data) || \int p(w, z, b\mid \theta, \Hb, \eta) p(\theta$, $\Hb\mid \data, \eta)) d\theta d\Hb$, or equivalently by minimizing,
\begin{equation}
\begin{aligned}[b]
J(&\phi_w, \phi_z, \phi_b  \mid \Hb^s, \theta^s)= -\E_{q(z)}[\sum_n\text{ln }p(\xseq \mid z_n, \theta^s)] \nonumber \\
& - \E_{q(z)q(b)q(w)}[\sum_n\text{ln }p(z_n \mid b_n, w_n, \Hb^s, \tau)] \\
& - \sum_n\ent{q_{\phi_z}(z_n\mid \xseq, y_n)} \nonumber \\
& + \sum_n\KL{q_{\phi_b}(b_n\mid \xseq, y_n)}{p(b_n \mid \gamma)} \\
& + \sum_n\KL{q_{\phi_w}(w_n\mid \xseq, y_n)}{p(w_n)}, 
\label{eq:vfe}
\end{aligned}[b]
\end{equation}
where $\ent{q}$ is the entropy of the approximation $q(z)$ and $\theta^s, \Hb^s \sim p(\theta$, $\Hb\mid \data, \eta)$.  
Putting it all together, our algorithm proceeds by cycling between a pSGLD step and an ADAM gradient step to minimize Equation~\ref{eq:vfe}, alternating between minimizing the posterior divergence of the local variables and sampling from the posterior of the global variables. We use standard reparameterized gradients~\cite{kingma2013auto} to handle the intractable expectations in Equation~\ref{eq:vfe}. Both steps are amenable to mini-batching and we use only a mini-batch of $\data$ in practice.  A sketch of the algorithm is available in the supplement.
Inference in the unconditional model \mname is simpler and only requires a few minor tweaks --- we only need to sample $\theta$ via SG-MCMC and since no meta-data $y$ is available, the variational approximation does not require a product-of-experts structure. 

We use SG-MCMC to sample the global variables. This requires the gradient of the marginal distribution $p(\data, \theta, \Hb \mid \eta)$. We use an importance sampling approximation to estimate, $\grad{\theta, \Hb} \text{ ln }p(\data, \theta, \Hb \mid \eta)$.
The steps in the inference algorithm are summarized in Algorithm 1. 
\begin{algorithm}[!h]
\caption{\cmname inference}
\label{algo:ehrvae}
\SetKwInOut{Input}{Input}
\SetKwInOut{Output}{Output}
\Input{$\data = \{\xseq\}_{n=1}^N, \{y_n\}_{n=1}^N$}
$\theta$, $\phi_w$, $\phi_z$, $\phi_b$ $\Hb$, $\eta$ $\gets$ Initialize parameters\\
\For{A fixed number of iterations}{
$\data_M = \{\xseq\}_{n=1}^M$ $\gets$ Random minibatch of M patients \\
For $n \in \data_M$, Sample local variables from their variational approximations, $w_n^s \sim q_{\phi_w}(w_n^s \mid x_n, y_n)$, $b_n^s\sim q_{\phi_b}(b_n^s \mid x_n, y_n)$, and $z_n^s \sim q_{\phi_z}(z_n^s \mid x_n, y_n)$\\
Use $w_n^s, b_n^s, x_n^s$ to form an one sample importance sampling estimate of the marginal $p(\Hb, \theta \mid \{\xseq\}_{n=1}^M, \eta)$\\
Sample $\theta^s, \Hb^s \sim p(\Hb, \theta \mid \{\xseq\}_{n=1}^M, \eta)$\\
Update $\phi_w, \phi_z, \phi_b$ $\gets$ ADAM(J($\phi_w, \phi_x, \phi_b \mid \Hb^s, \theta^s$))
}
\Output{ $\theta$, $\phi_w, \phi_z, \phi_b$, $\Hb$}
\label{alg:inf1}
\end{algorithm}
\section{Related work}
\textbf{Generative Models}
have seen a resurgence in interest of late. Latent variable based deep directed models --- Generative adversarial networks (GAN)~\cite{goodfellow2014generative}, variational autoencoders~\cite{kingma2013auto} and their variants have been effective at generating a wide variety of content from natural images to chemical structures~\cite{gomez2016automatic}. While similar, there are important distinctions between the two model classes. VAEs specify an explicit parametric distribution over observations, GANs, on the other hand, are \emph{likelihood free} and define a stochastic procedure for directly generating the data. Learning in GANs proceeds by comparing the generated data with real data and backpropagating gradients to guide the stochastic data generating procedure. Non-differentiability induced by discrete data make a direct application of GANs intractable. While a few adaptations of GANs to discrete data have been attempted~\cite{yu2017seqgan}, by and large the problem remains challenging. 
In contrast, generating non-continuous data in the VAE framework is tractable as long as an appropriate discrete density can be specified for the data. Autoregressive distributions, when suitably defined, are able to account for correlations exhibited by the data and prove convenient for specifying flexible densities over spatio-temporal sequences~\cite{salimans2017pixelcnn++}. VAE models combined with such autoregressive densities have shown promise in generating discrete text fragments~\cite{bowman2015generating, hu2017toward, yang2017improved}. Our models, \mname and conditional \mname, are both examples of this category. 

\textbf{Conditional variants} of VAEs~\cite{kingma2014semi, NSid17} have previously been explored to learn from limited labeled data~\cite{kingma2014semi} and recover disentangled representations~\cite{NSid17} 
Our work extends these models by introducing hierarchically factorized latent variables, with the upper level of the hierarchy shared across the population. By explicitly disentangling factors of variations stemming from medical conditions from those arising from individual differences among patients, the representations learned by our models are easier to intuit. Controlled generation has also been explored in~\cite{hu2017toward}, unlike us they do not attempt to infer population wide latent variables and have to rely on continuous relaxation to discrete data. Others have explored hierarchical VAEs~\cite{hsu2017unsupervised}, but they are non-conditional and unable to exploit available meta-data. Also tangentially related are temporal extensions of VAEs that endow each time step of a sequence with its own latent variable~\cite{VRNN_Chung}. Such models primarily care about modeling the temporal dynamics of the sequences but do not attempt to recover sequence level or population level representations.

\textbf{Synthetic EHR generation} In spite of the widespread adoption of EHR systems by health care systems, this data remains largely siloed owing to patient privacy concerns.  
Synthetically generated EHRs hold the promise of alleviating such issues and have received some attention in the past. Systems that rely on hand-engineered rules~\cite{walonoski2017synthea, mclachlan2018aten} and that are tailored to specific disease conditions~\cite{Buczak2010} have previously been explored. However, these tend to be brittle and are difficult to generalize beyond the particular diseases considered while developing the system. 
More closely related to our work are recent efforts that leverage deep generative models for synthesizing EHRs~\citep{choi2017generating, beaulieu2017privacy, baowaly2018synthesizing}. While promising, these approaches are limited by their inability to generate sequences, instead only generating a single patient representation aggregated over time. They thus lose important temporal characteristics of real world longitudinal EHRs. Our work extends this line of work by generating sequential health records.



\section{Experiments}\label{sec:exp}
In this section, we vet the proposed methods along different aspects: 
\begin{itemize}[leftmargin=*]
    \item {\bf Capturing EHR statistics:} the degree to which they model real world EHRs,
    \item {\bf Usefulness of synthetic EHR:} their ability to generate realistic and useful synthetic data, 
    \item {\bf Privacy:} assessment of the privacy preservation.
\end{itemize}

We begin by describing our experimental setup and data and then proceed to describing the experiments and baselines. 

\paragraph{\textbf{Source data}} The data used in this study was sourced from a large medical  center. It consists of 10-years of longitudinal medical records of total 258,555 patients with 207,384 training and 51,171 testing amounting to over 13 million visits. This  dataset was cleaned and preprocessed to obtain patient level sequences. A detailed description of the preprocessing as well as summary statistics describing the data are available in the supplement. 

\paragraph{\textbf{Methods for comparison}} We compare several approaches proposed for modeling sequential data:

\begin{itemize}[leftmargin=*]
    \item LSTM: Our first baseline consists of a language LSTM model. For a controlled comparison, we use bi-directional LSTM networks to parametrize the inference network for all models.
    \item VAE-LSTM: We also compare against VAE based models that have been used for modeling discrete sequences. This includes, VAE with an LSTM decoder (VAE-LSTM)~\cite{bowman2015generating}
    \item VAE-Deconv: We replace LSTM with a deconvolution network to develop a variant called VAE-Deconv \cite{semeniuta2017hybrid}.
    \item \mname: This model is our proposed model (Section~\ref{sec:eva}).
    \item \cmname: This is the conditional variant of \mname (Section~\ref{sec:evac}).
\end{itemize}


In all subsequent experiments, we split the real data into a 80/20 train/test split. We train the various generative models only on the training split holding out the remaining for evaluation. For \cmname we model the ten most prevalent conditions in the real dataset and lump all other conditions into a ``background condition''. 
%
%
\begin{figure*}[ht]
\centering
\includegraphics[width=1\textwidth]{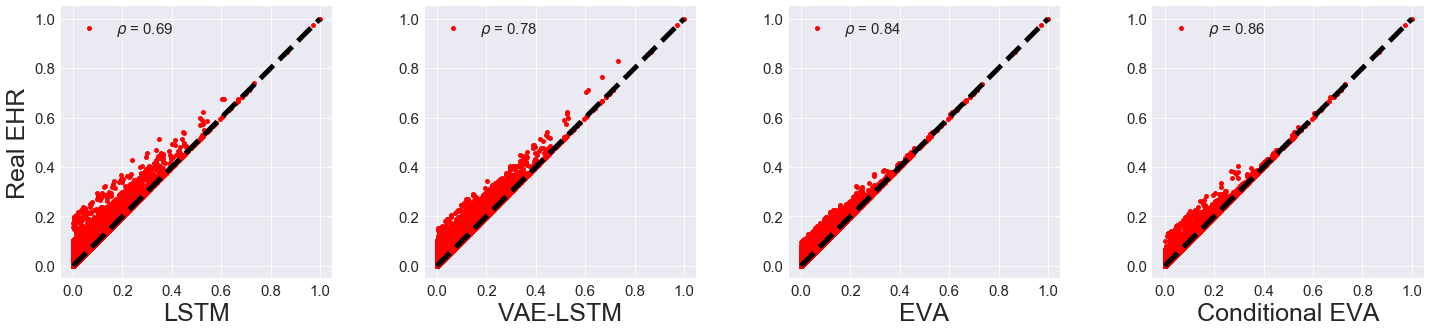}
  \caption{\small{Comparison of bigram statitics of generated and real EHR codes. It confirmed that \mname generates data that capture better correlations in the EHR data.}}
\label{fig:bigram_scatter}
\end{figure*}

\subsection{Capturing EHR statistics}
We present evaluations of how well the proposed methods capture EHR statistics in Figure~\ref{fig:pred_check}, which plots the predictive log-likelihoods achieved by the different methods on the PAMF test set. Our models \mname and \cmname produce higher test log-likelihoods, which suggests that the proposed models better capture statistical regularities of EHRs. 
 After training on real dataset, we generated synthetic EHR sequences, equal in number to the number of training sequences. We then calculated the marginal probability of occurrence of bi-gram tokens. Figure~\ref{fig:bigram_scatter} provides a comparison of these statistics between the real and synthetic data through a scatter plot and pair-wise Pearson correlation coefficients ($\rho$). More comparisons are available in the supplement. It is easy to see that our models provide dramatic improvements over competing approaches with much higher Pearson correlation. 

%

%
%
\subsection{Usefulness of synthetic EHRs}
Classification models are often developed on EHR data to predict whether a patient will develop a certain disease. To evaluate the utility of our synthetic EHRs, we tested how well such prediction tasks are supported by the synthetic longitudinal EHR data. To do so, we set up a task for predicting $P(x_{n, t} \mid x_{n, t-1}, \ldots, x_{n, 1})$--- a patient's future condition given her history. We generated $258,555$ synthetic patient sequences from the different generative models trained on an identical number of real patients. We held-out $20\%$ of the data for testing and trained a Long short-term memory network(LSTM) model on the remainder. We repeated this process five times each with a separate random split. We also repeated the process for real data to obtain an estimate of the upper-bound on performance. 

\begin{figure*}[ht]
\centering
\includegraphics[width=1\textwidth]{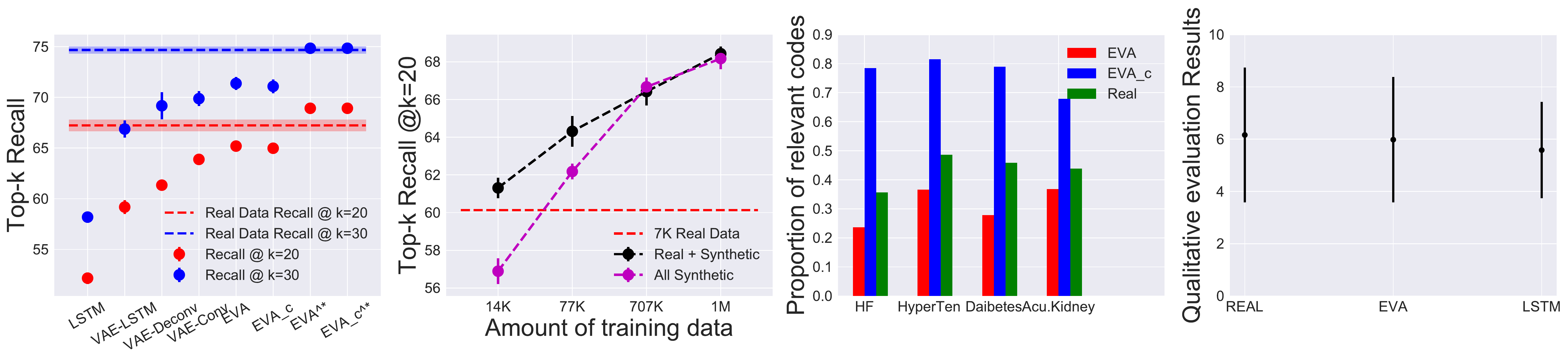}
  \caption{\small{
 1) displays the average test log likelihoods; 2) top-20 recall performance of \mname as a function of the number of synthetic records, 3) a comparison of proportion specific codes generated by \mname and \cmname; 4)  User study results of a clinician, rated on a 10 point scale, with 10 being most realistic and 1 being least realistic. 
 The error bars indicate two standard deviations. Both \mname and conditional \mname improve significantly over the competition both in terms of held-out log likelihood, matching marginal statistics of real EHRs and improved predictive performance. \cmname is able to generate condition-specific EHRs.}}
\label{fig:pred_check}
\end{figure*}

\noindent{\bf Accurate temporal prediction:} Following~\cite{choi2016doctor} we measure the performance in predicting $x_{n,t}$ using
the top-k recall metric. This is defined as the ratio between the number of true positives in the top k predictions and the total number of true positives. Again following~\cite{choi2016doctor}, we evaluate the different models at $k=20$ and $k=30$. Figure \ref{fig:pred_check} (1) summarizes the results. It can be seen that all VAE-based models significantly improve upon the sequence to sequence LSTM baseline. Moreover, \mname and \cmname are better than the competing VAE models and are closest to real data performance. Finally, we note that an LSTM trained on three million patients generated from \mname and \cmname \emph{outperforms variants trained on (smaller) real data}. 

\noindent{\bf Beating Data via Data Augmentation:} To further investigate whether augmenting limited real EHRs with synthetic ones is beneficial, we selected, uniformly at random, a $7000$ patient cohort from the test split of the real  dataset. We then augmented this data with varying amounts of synthetically generated data and repeated the predictive experiments. The results are shown in Figure~\ref{fig:pred_check} (2). Here, it can be seen that when we use only synthetic data, we need about an order of magnitude larger dataset to exceed real data performance. However, augmenting the $7$K real patients with only an additional $7$K synthetic records already outperforms the real data. Finally, when the amount of synthetic data grows, the effect of augmentation wanes. This is because the resulting dataset is dominated by the synthetic data. At $1M$, the synthetic records outperform the real data by about eight percent.  

\begin{figure*}[t]
\centering
\includegraphics[width=0.3\textwidth]{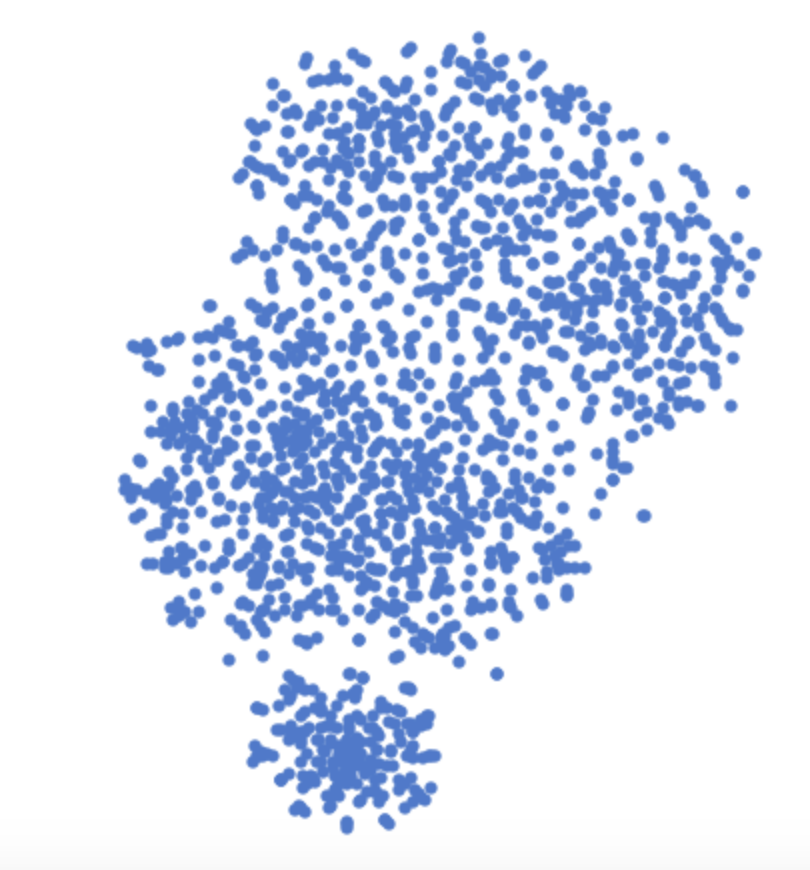}
\includegraphics[width=0.3\textwidth]{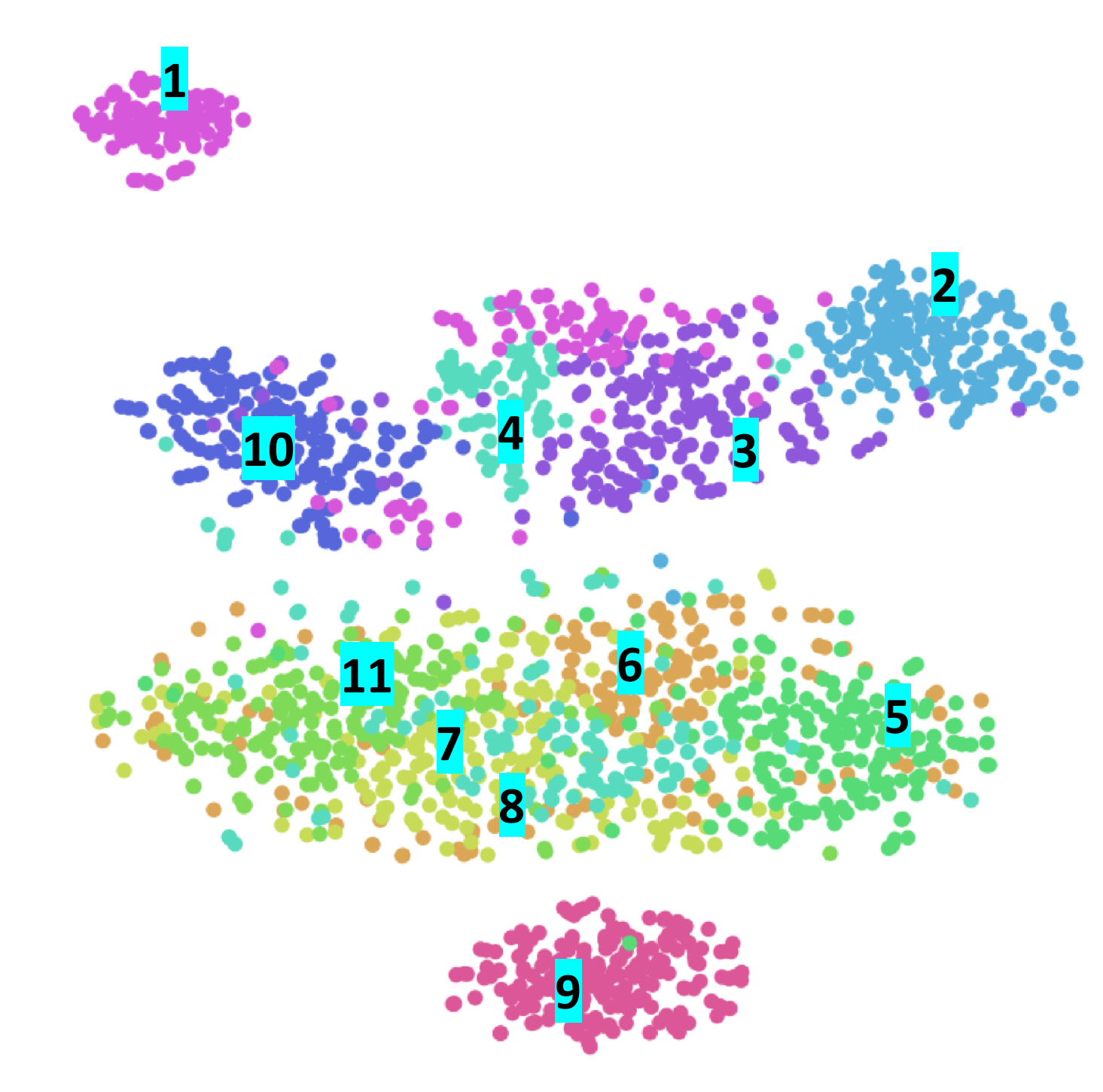}
\caption{\small{tSNE plots to illustrate latent space learned by \mname and \cmname (right). As we note the clusters learned by \mname (left) have no clear separation,  while  the clusters present in embedding space of \cmname indicate a disentangled representation of the latent space. \cmname clusters are binned by different conditions. The numbers indicate the following disease conditions, 1: Diabetes mellitus , 2: Cancer of brain and nervous system, 
3: Epilepsy, convulsions, 
4: Chronic kidney disease, 5: Cough/Cold, 6: Fever, 7: Viral infection, 8: non-epithelial cancer of skin, 9: heart failure, 10: Hypertension with complications and secondary hypertension, and 11: Background with none of the above conditions. }}
\label{fig:tSNE}
\end{figure*}
\subsection{Benefits of conditional generation} Thus far, we have seen that \cmname and \mname perform similarly in generating realistic, but generic, EHR sequences. The conditional model; however, is more interesting because of its ability to generate condition specific EHRs. Figure~\ref{fig:pred_check}(3) contrasts the statistics of records generated by \mname against those generated by \cmname conditioned on four common conditions, heart failure, essential hypertension, acute kidney failure, and type 2 diabetes mellitus. From the large proportion of condition specific codes, it is evident that \cmname is effective at controlled EHR generation. To further test whether condition specific EHR generated by \cmname can outperform generic EHRs generated by \mname, we considered the problem of predicting whether a patient will have an heart failure given her history. To do this, we need a dataset of cases, patients with heart failure, and controls, patients without heart failure. The real dataset contains $3800$ heart failure cases and $280,000$ controls. We used \cmname to match these numbers and generated cases by conditioning on the heart failure condition and controls by conditioning on the background condition. An LSTM trained on this data produced an area under the ROC curve (AUC) score of $\mathbf{74.66}$, which is comparable to the real data score of $\mathbf{76.75}$. Generation of such data with cases and controls cannot be achieved using \mname because it can only produce generic EHR sequences, but the heart failure-specific sequences generated by \cmname result in improved performance. Figure~\ref{fig:tSNE} provides a visualization of the latent space recovered by the two models. While the representations produced by \cmname cluster with respect to the medical conditions, no such clustering is observed for \mname.

\subsection{Clinical User Study}
 We also performed qualitative evaluation of the synthetic EHR  sequences by recruiting clinical experts for a user study
We presented clinicians with fifty real and fifty \mname generated synthetic patient records selected at random. The clinician was not made aware of whether a record was real or synthetic, and asked to judge whether the record seemed realistic. Records were rated on a ten point scale with one being least realistic. Figure~\ref{fig:pred_check}(4) provides the results of this analysis. We find that clinicians find \mname generated records just as realistic as real patient records. Although the average score is just around 6 out 10, it is mainly due to the details associated with EHR structured data are not available, which raise the importance of further research in generating more detailed data such as clinical notes in future. 

\subsection{Diversity within visits}
 As a measure of high quality EHR data generation algorithm, \mname should be able to generate diverse yet relevant patient visits sequences. To evaluate the diversity of visits within a single patient in our data, we measured Jaccard Similarity index between visits. We calculate Jaccard similarity index from one visit to next visit and finally average over the entire sequence. The lower Jaccard Similarity index, the more diverse result is. The average Jaccard similarity index for real EHR is 0.1835 and for synthetic EHR data generated by \mname and \cmname are 0.2167 and 0.2235, which are close to the real EHR. Comparatively other baselines achieved much higher average Jaccard similarity index as shown in Table \ref{tab:Jaccard_coeff}.
 
\begin{table}[]
\caption{Average Jaccard Similarity Index}
\begin{tabular}{l|l}
Model          & Jaccard Coefficient \\ \hline
LSTM       & 0.3874              \\ 
VAE-LSTM   & 0.3156              \\ 
VAE-Deconv & 0.2631              \\ \hline
\mname        & 0.2167              \\ 
\cmname       & 0.2235              \\ \hline
Real EHR   & 0.1835              \\ \hline
\end{tabular}
\label{tab:Jaccard_coeff}
\end{table}
 
\noindent{\bf Weight uncertainty in $\theta$}: We find that inferring a distribution over $\theta$, rather than a point estimate, is crucial for the generation of diverse and hence realistic sequences. The point estimate variants  produce sequences with unnaturally many repeated tokens in a sequence. To quantitatively evaluate this effect, we generated 250k synthetic EHRs from EVA and its point estimate variant. For each sequence we computed the ratio of the number of unique tokens to the total number of tokens in the sequence, which we averaged over all sequences. The \mname
point estimate produced a score of $0.2786 \pm 0.08$, while the Bayesian variant scored $0.3214 \pm 0.07$ and real EHRs exhibit a ratio of $0.3845 \pm 0.08$. We also present some sample patient data illustrating generated EHR sequences in Table~\ref{table:EVA}.

\subsection{Privacy Risk Evaluation}
EHR de-identification or generation tasks always have the privacy risk where there is often way to retrieve the underlying original records. While intuitively by training \mname on training data and generating samples overcomes 1-to-1 mapping from original data to generated data, we wanted to formally evaluate privacy preserving aspects of the generated data. In this section we have performed a formal assessment of \mname’s privacy risks.

\begin{figure}[t]
\centering
    \begin{subfigure}[t]{0.26\textwidth}
        \includegraphics[height=1.5 in]{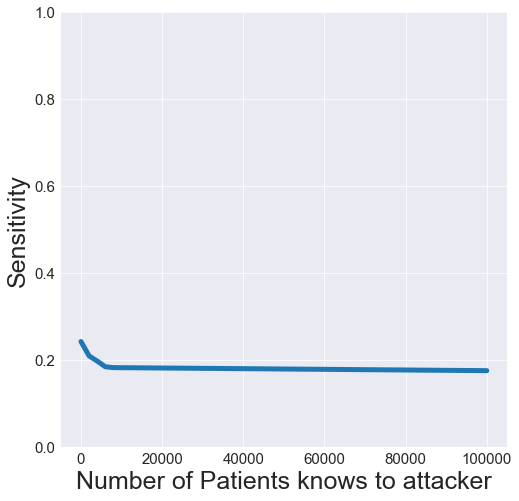}
        \caption{Sensitivity}
        \label{fig:sensitivity}
    \end{subfigure}%
    \begin{subfigure}[t]{0.26\textwidth}
        \includegraphics[height=1.5 in]{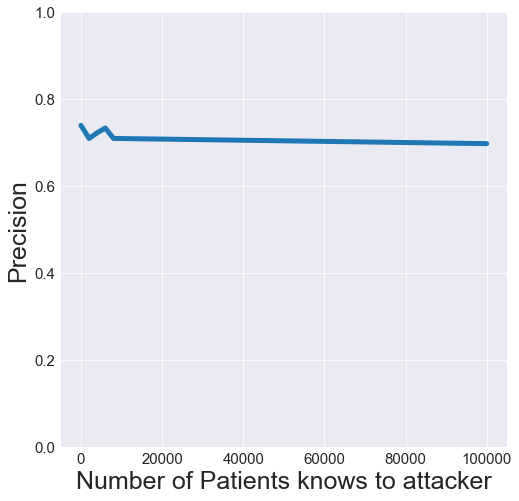}
        \caption{Precision}
         \label{fig:precision}
    \end{subfigure}

    \caption{ Sensitivity and precision vs. number of patients known to the attacker. 
}
    \label{fig:privacy}
\end{figure}

\textbf{Presence disclosure} occurs when an attacker can determine that \mname was trained with a dataset including the record from patient x. We assume the attacker will check if any synthetic record matches records from x by ignoring the visit orders\footnote{If the visit order is enforced in matching, the matching probability will decrease hence the attack success probability.}. If matches, the attacker will assume x is in the training data.   
We assume a prior probability 0.8 that x is in the training data. In practice, the prior probability can be much lower in that case the successful attack will be much harder. We use sensivity and precision as metrics of attack success.
Figure \ref{fig:privacy} depicts the sensitivity (i.e. recall) and the precision of the presence disclosure test when varying the number of real patient the attacker knows. In this case, x\% sensitivity means the attacker has successfully discovered that x\% of the records that he/she already knows were used to train \mname. Similarly, x\% precision means, when an attacker claims that a certain number of patients were used for training \mname, only x\% of them were actually used. Since the prior success probability is 0.8, we want to assess whether the attacker can gain additional knowledge by improving sensitivity and precision above 0.8. 

Figure ~\ref{fig:sensitivity} shows that with attacker can only discover 20\% percent of the known patients to attacker were used to train \mname, which is much lower than a prior 80\%. Similarly Figure~\ref{fig:precision} shows that, the precision is around 70\% which is again lower than the prior precision 80\%. In fact, this indicates that by analyzing the synthetic data the attacker do not gain any additional knowledge to improve their success probability (the precision and sensitivity actually reduced). 
This confirmed that the synthetic data are not useful  for presence disclosure attack. Note that how many real patients known to the attacker do not change the attacker's performance.

\subsection{Sample Data and Generated Data}
We present two sample patient records generated by  RNN and \mname  in this table \ref{table:RNN} and \ref{table:EVA}, respectively. Here we presented diagnosis(D), Rx/medication(R), Procedure(P) codes of two patients across 6 clinical visits. We observe that the records generated by \mname are considerable more diverse in terms of clinical events than the one generated by LSTM. In fact, the ones generated by LSTM has large percentage of repeated events across visits than the ones generated by \mname. This provides an intuitive demonstration of the effectiveness of \mname in generating diverse records.

\begin{table}[ht]
\caption{Example records generated by LSTM}
\scalebox{0.8}{
\begin{tabular}{ll}
    \hline
    \multirow{3}{*}{visit-1}&D:Other screening for suspected conditions\\ &P: Pathology\\ & D: Cardiac dysrhythmias        \\
    \hline
    \multirow{1}{*}{visit-2}&D: Medical,examination/evaluation          \\
    \hline
    \multirow{2}{*}{visit-3}&R: Beta blockers  \\ &P: Pathology        \\
    \hline
    \multirow{4}{*}{visit-4}&D: Other screening for suspected   \\  &D:conditions not mental disorders or infectious disease  \\ & R: Beta blockers 
    \\ & D: Cardiac dysrhythmias
    \\
    \multirow{2}{*}{visit-5}&R: Beta blockers   \\ &D: Cardiac dysrhythmias     \\
    \hline
    \multirow{2}{*}{visit-6}&R: Beta blockers   \\ &R: Beta blockers        \\
    \hline
\end{tabular}
}
\label{table:RNN}
\end{table}

\begin{table}[ht]
\caption{Example records generated by \mname}
\scalebox{0.8}{
\begin{tabular}{ll}
    \hline
    \multirow{5}{*}{visit-1}& D:Disorders of lipid metabolism\\ &D:Immunizations and screening for infectious disease\\ & D:Other liver disease 
    \\& D: Peripheral and visceral atherosclerosis
    \\
    \hline
    \multirow{4}{*}{visit-2}&R Antidepressants           \\&R Antihypertensive        \\&D: Hyperplasia of prostate        \\&D: Mood disorders             \\
    \hline
    \multirow{4}{*}{visit-3}&R: Antihyperlipidemic   \\ &D Coronary atherosclerosis and other heart disease   \\&R: Beta blockers    \\&R: Antihypertensive  \\
    \hline
    \multirow{5}{*}{visit-4}&R: Misc. Hematological \\  &D: Disorders of lipid metabolism \\ & D: Hyperplasia of prostate  
    \\ & D: Coronary atherosclerosis and other heart disease \\ & D: Mood disorders   
    \\
    \hline
    \multirow{6}{*}{visit-5}&D: Disorders of lipid metabolism  \\ &D: Other liver diseases  \\ &D: Coronary atherosclerosis and other heart disease\\ &D: Peripheral and visceral atherosclerosis   \\ &D: Other upper respiratory disease\\ &D: Medical examination/evaluation  \\
    \hline
    \multirow{2}{*}{visit-6}&R: Hypnotics   \\ &D: Mood disorders   \\
    \hline
\end{tabular}
}
\label{table:EVA}
\end{table}

\section{Discussion \& Conclusions}\label{sec:concl}
The findings from this study suggest that synthetic, but realistic, longitudinal EHR data can be generated.  This is notable because it can enable the dissemination of privacy-respective data for research and development of artificial intelligence applications in healthcare. 
Our experimental results further show that synthetic sequences of EHR data can be used as a drop-in replacement for real data without significantly sacrificing performance for building sequence models, such as one might do with a recurrent neural network.
A combination of real and synthetic data outperforms either in isolation. Moreover, a user study with a clinician confirmed that the generated EHR sequences are realistic.

Still, there are opportunities for advancement in the generation of synthetic longitudinal discrete data. In particular, though our models are adept at generating EHR sequences, they make no attempt at modeling time gaps between visits. Modeling such gaps is challenging because they are influenced by factors beyond physiology, including social- and economic-determinants of health and the healthcare insurance plan of the patient. 
Future investigations could  addressing this by incorporating additional data sources and modeling the inter-arrival time of clinical visits.
\bibliographystyle{unsrt}
\bibliography{sigproc}

\begin{thebibliography}{10}

\bibitem{henry2016adoption}
J~Henry, Yuriy Pylypchuk, and Vaishali Patel.
\newblock Adoption of electronic health record systems among us non-federal
  acute care hospitals: 2008-2015.
\newblock In {\em The {O}ffice of {N}ational {C}oordinator for {H}ealth
  {I}nformation {T}echnology, Data Brief.}, 2016.

\bibitem{musen2014clinical}
Mark~A Musen, Blackford Middleton, and Robert~A Greenes.
\newblock Clinical decision-support systems.
\newblock In {\em Biomedical Informatics}, pages 643--674. Springer, 2014.

\bibitem{choi2016doctor}
Edward Choi, Mohammad~Taha Bahadori, Andy Schuetz, Walter~F. Stewart, and
  Jimeng Sun.
\newblock Doctor {AI}: Predicting clinical events via recurrent neural
  networks.
\newblock In {\em MLHC}, 2016.

\bibitem{futoma2017improved}
Joseph Futoma, Sanjay Hariharan, Katherine Heller, Mark Sendak, Nathan Brajer,
  Meredith Clement, Armando Bedoya, and Cara O’Brien.
\newblock An improved multi-output gaussian process rnn with real-time
  validation for early sepsis detection.
\newblock In {\em MLHC}, pages 243--254, 2017.

\bibitem{JHO14}
Joyce~C. Ho, Joydeep Ghosh, and Jimeng Sun.
\newblock Marble: high-throughput phenotyping from electronic health records
  via sparse nonnegative tensor factorization.
\newblock In {\em KDD}, pages 115--124, 2014.

\bibitem{YWang15}
Yichen Wang, Robert Chen, Joydeep Ghosh, Joshua~C. Denny, Abel~N. Kho, You
  Chen, Bradley~A. Malin, and Jimeng Sun.
\newblock Rubik: Knowledge guided tensor factorization and completion for
  health data analytics.
\newblock In {\em KDD}, pages 1265--1274, 2015.

\bibitem{el2011systematic}
Khaled El~Emam, Elizabeth Jonker, Luk Arbuckle, and Bradley Malin.
\newblock A systematic review of re-identification attacks on health data.
\newblock {\em PloS one}, 6(12):e28071, 2011.

\bibitem{el2015anonymising}
Khaled El~Emam, Sam Rodgers, and Bradley Malin.
\newblock Anonymising and sharing individual patient data.
\newblock {\em bmj}, 350:h1139, 2015.

\bibitem{choi2017generating}
Edward Choi, Siddharth Biswal, Bradley Malin, Jon Duke, Walter~F Stewart, and
  Jimeng Sun.
\newblock Generating multi-label discrete patient records using generative
  adversarial networks.
\newblock In {\em MLHC}, 2017.

\bibitem{beaulieu2017privacy}
Brett~K Beaulieu-Jones, Zhiwei~Steven Wu, Chris Williams, and Casey~S Greene.
\newblock Privacy-preserving generative deep neural networks support clinical
  data sharing.
\newblock {\em bioRxiv}, page 159756, 2017.

\bibitem{baowaly2018synthesizing}
Mrinal~Kanti Baowaly, Chia-Ching Lin, Chao-Lin Liu, and Kuan-Ta Chen.
\newblock Synthesizing electronic health records using improved generative
  adversarial networks.
\newblock {\em Journal of the American Medical Informatics Association}, 2018.

\bibitem{goodfellow2014generative}
Ian Goodfellow, Jean Pouget-Abadie, Mehdi Mirza, Bing Xu, David Warde-Farley,
  Sherjil Ozair, Aaron Courville, and Yoshua Bengio.
\newblock Generative adversarial nets.
\newblock In {\em NIPS}, pages 2672--2680, 2014.

\bibitem{van2016wavenet}
Aaron Van Den~Oord, Sander Dieleman, Heiga Zen, Karen Simonyan, Oriol Vinyals,
  Alex Graves, Nal Kalchbrenner, Andrew Senior, and Koray Kavukcuoglu.
\newblock Wavenet: A generative model for raw audio.
\newblock {\em arXiv preprint arXiv:1609.03499}, 2016.

\bibitem{bowman2015generating}
Samuel~R Bowman, Luke Vilnis, Oriol Vinyals, Andrew~M Dai, Rafal Jozefowicz,
  and Samy Bengio.
\newblock Generating sentences from a continuous space.
\newblock {\em arXiv preprint arXiv:1511.06349}, 2015.

\bibitem{kingma2013auto}
Diederik~P Kingma and Max Welling.
\newblock Auto-encoding variational bayes.
\newblock {\em arXiv preprint arXiv:1312.6114}, 2013.

\bibitem{DAckley1987}
David~H Ackley, Geoffrey~E Hinton, and Terrence~J Sejnowski.
\newblock A learning algorithm for boltzmann machines.
\newblock In {\em Readings in Computer Vision}, pages 522--533. Elsevier, 1987.

\bibitem{miao2016language}
Yishu Miao and Phil Blunsom.
\newblock Language as a latent variable: Discrete generative models for
  sentence compression.
\newblock In {\em EMNLP}, 2016.

\bibitem{yang2017improved}
Zichao Yang, Zhiting Hu, Ruslan Salakhutdinov, and Taylor Berg-Kirkpatrick.
\newblock Improved variational autoencoders for text modeling using dilated
  convolutions.
\newblock In {\em International Conference on Machine Learning}, pages
  3881--3890, 2017.

\bibitem{semeniuta2017hybrid}
Stanislau Semeniuta, Aliaksei Severyn, and Erhardt Barth.
\newblock A hybrid convolutional variational autoencoder for text generation.
\newblock {\em arXiv preprint arXiv:1702.02390}, 2017.

\bibitem{kingma2014semi}
Diederik~P Kingma, Shakir Mohamed, Danilo~Jimenez Rezende, and Max Welling.
\newblock Semi-supervised learning with deep generative models.
\newblock In {\em Advances in Neural Information Processing Systems}, pages
  3581--3589, 2014.

\bibitem{gershman2014amortized}
Samuel Gershman and Noah Goodman.
\newblock Amortized inference in probabilistic reasoning.
\newblock In {\em {P}roceedings of the {A}nnual {M}eeting of the {C}ognitive
  {S}cience {S}ociety}, volume~36, 2014.

\bibitem{wu2018multimodal}
Mike Wu and Noah Goodman.
\newblock Multimodal generative models for scalable weakly-supervised learning.
\newblock In {\em {I}nternational {C}onference on {M}achine {L}earning}, 2018.

\bibitem{li2016preconditioned}
Chunyuan Li, Changyou Chen, David~E Carlson, and Lawrence Carin.
\newblock Preconditioned stochastic gradient langevin dynamics for deep neural
  networks.
\newblock In {\em AAAI}, volume~2, page~4, 2016.

\bibitem{gomez2016automatic}
Rafael G{\'o}mez-Bombarelli, Jennifer~N Wei, David Duvenaud, Jos{\'e}~Miguel
  Hern{\'a}ndez-Lobato, Benjam{\'\i}n S{\'a}nchez-Lengeling, Dennis Sheberla,
  Jorge Aguilera-Iparraguirre, Timothy~D Hirzel, Ryan~P Adams, and Al{\'a}n
  Aspuru-Guzik.
\newblock Automatic chemical design using a data-driven continuous
  representation of molecules.
\newblock {\em ACS Central Science}, 2016.

\bibitem{yu2017seqgan}
Lantao Yu, Weinan Zhang, Jun Wang, and Yong Yu.
\newblock Seqgan: Sequence generative adversarial nets with policy gradient.
\newblock In {\em AAAI}, pages 2852--2858, 2017.

\bibitem{salimans2017pixelcnn++}
Tim Salimans, Andrej Karpathy, Xi~Chen, and Diederik~P Kingma.
\newblock Pixelcnn++: Improving the pixelcnn with discretized logistic mixture
  likelihood and other modifications.
\newblock {\em arXiv preprint arXiv:1701.05517}, 2017.

\bibitem{hu2017toward}
Zhiting Hu, Zichao Yang, Xiaodan Liang, Ruslan Salakhutdinov, and Eric~P Xing.
\newblock Toward controlled generation of text.
\newblock In {\em International Conference on Machine Learning}, pages
  1587--1596, 2017.

\bibitem{NSid17}
Siddharth Narayanaswamy, T.~Brooks Paige, Jan-Willem van~de Meent, Alban
  Desmaison, Noah Goodman, Pushmeet Kohli, Frank Wood, and Philip Torr.
\newblock Learning disentangled representations with semi-supervised deep
  generative models.
\newblock In {\em NIPS}, pages 5925--5935. 2017.

\bibitem{hsu2017unsupervised}
Wei-Ning Hsu, Yu~Zhang, and James Glass.
\newblock Unsupervised learning of disentangled and interpretable
  representations from sequential data.
\newblock In {\em NIPS}, pages 1876--1887, 2017.

\bibitem{VRNN_Chung}
Junyoung Chung, Kyle Kastner, Laurent Dinh, Kratarth Goel, Aaron~C. Courville,
  and Yoshua Bengio.
\newblock A recurrent latent variable model for sequential data.
\newblock {\em CoRR}, abs/1506.02216, 2015.

\bibitem{walonoski2017synthea}
Jason Walonoski, Mark Kramer, Joseph Nichols, Andre Quina, Chris Moesel, Dylan
  Hall, Carlton Duffett, Kudakwashe Dube, Thomas Gallagher, and Scott
  McLachlan.
\newblock Synthea: An approach, method, and software mechanism for generating
  synthetic patients and the synthetic electronic health care record.
\newblock {\em JAMIA}, page ocx079, 2017.

\bibitem{mclachlan2018aten}
Scott McLachlan, Kudakwashe Dube, Thomas Gallagher, Bridget Daley, and Jason
  Walonoski.
\newblock The aten framework for creating the realistic synthetic electronic
  health record.
\newblock 2018.

\bibitem{Buczak2010}
Anna Buczak, Steven Babin, and Linda Moniz.
\newblock Data-driven approach for creating synthetic electronic medical
  records.
\newblock {\em BMC Medical Informatics and Decision Making}, 10(1):59, 2010.

\bibitem{kingma2014adam}
Diederik~P Kingma and Jimmy Ba.
\newblock Adam: A method for stochastic optimization.
\newblock {\em arXiv preprint arXiv:1412.6980}, 2014.

\end{thebibliography}
\clearpage
\newpage
\section{Supplement}
\subsection{EHR Preprocessing}
In this section, we describe the preprocessing procedure used by our models.  We note that we also experimented with a variant that does away with this pre-processing and instead models each visit $x_{n,t}$ as a collection of independent Bernoulli distributions. However, we had trouble learning this variant and it performed significantly worse in preliminary evaluations. 
This preprocessing step is also described in algorithm \ref{algo:preprocessing1}.

\paragraph{\textbf{Creation of vocabulary}}: Since longitudinal EHR data can represented as sequence of visits where each visit consist of few different ICD-9 codes, we combine the ICD-9 codes into a visit representation. For example a patient can be represented as this [[$c_1$, $c_2$, $c_3$], [$c_{10}$, $c_{34}$], [$c_{21}$,$c_{34}$]], where [$c_1$,$c_2$,$c_3$] is a visit. We first identify all the unique visits combinations such as this [$c_1$,$c_2$,$c_3$] and calculate the frequency of all these visit combinations. We select top 50,000 most frequent visit combination as our vocabulary. 

\paragraph{\textbf{Replacement of less frequent visit combination}}: In order to be train the algorithm, we first decided to use a vocabulary of size 50k. Since the total number of unique combination of visits is around 2 million but many of them only appear once in the entire EHR dataset, we decided to replace those visit combinations with visit combination from top 50k visit combination. The replacement was done by finding the best matching intersection between the sets in top 50k vocabulary with rest of the set in the unique combination sets.  

\begin{algorithm}[ht]
\caption{EHR preprocessing}
\label{algo:preprocessing1}
\SetKwInput{Input}{Input}
\SetKwInput{Output}{Output}
\DontPrintSemicolon
\vspace{0.1cm}
\Indm
\Input{Longitudinal EHR dataset  $D = \{P_1,\dots,P_N\}$ and Patient $P_i$ is combination of visits $V_i$}
\Output{Modified EHR Dataset $D$ where   $D = \{P_1,\dots,P_N\}$ and Patient $P_i$ is combination of visits in Visit Vocab $W$}
\Indp
\vspace{0.1cm}
Unique visits UV $\gets$   Collect all unique visits combination in D \\
Visit Vocab W $\gets$ Sort visits by frequency of occurrence in UV and select top 50,000 visits \\

Replacement Dictionary RD $\gets$ Find replacement of visits not in Visit Vocab W by finding closest matching Visit in Visit Vocab W

\ForEach{$P_i$ in $[P_1,\dots,P_N]$}{
	\ForEach{$V_i$ in $P_i$}{
      \eIf{$V_i$ in $W$}{
         keep $V_i$\;
         }{
         replace $V_i$ by finding replacement from RD \;
        }
    }
}
\end{algorithm}

\subsection{Additional Experiments and Data}
We begin this section by summarizing the datasets used.
\begin{table}[!h]
\caption{Basic statistics of dataset used}
\label{tab:data_stats}
\centering
\begin{tabular}{@{}l|c@{}}
\textbf{Dataset} & \textbf{Real dataset} \\
\toprule
\# of patients & 258,555  \\ 
\# of visits & 13,920,759  \\
Avg. \# of visits per patient & 53.8 \\
\# of unique ICD9 codes & 10,437 \\
Avg. \# of codes per visit & 1.98  \\
Max \# of codes per visit & 54 \\
\bottomrule
\end{tabular}
\vspace{-3mm}
\end{table}
\paragraph{Implementation Details}
For training models, we used Adam \citep{kingma2014adam} with a batch size of 32 samples, on a machine equipped with Intel Xeon E5-2640, 256GB RAM, eight Nvidia Titan-X GPU and CUDA 8.0. 

\noindent {\textbf{Hyperparameter Tuning: }}We define five hyper-parameters for \mname:
\vspace*{-1mm}
\begin{itemize}[leftmargin=5.5mm]
\item learning rate $\eb_i$: [ 2e-3,1e-3, 7e-4]
\item dimensionality $r$ of the LSTM hidden layer $\hb_t$ from Eq.: [100, 200, 300, 400, 500]
\item dropout rate for the dropout on the LSTM hidden layer: [0.0, 0.2, 0.4, 0.6, 0.8]
\item convolution kernels [3,5,7,10,15]
\item dilation kernels  [1, 2, 4, 8, 16]
\end{itemize}
The hyperparameters used in the final model were searched using random search. In order to fairly compare the model performances, we matched the number of model parameters to be similar for all baseline methods.


\subsection{Scatter Plots for uni-gram statistics}
\begin{figure*}[t]
\centering
\includegraphics[width=0.9\textwidth]{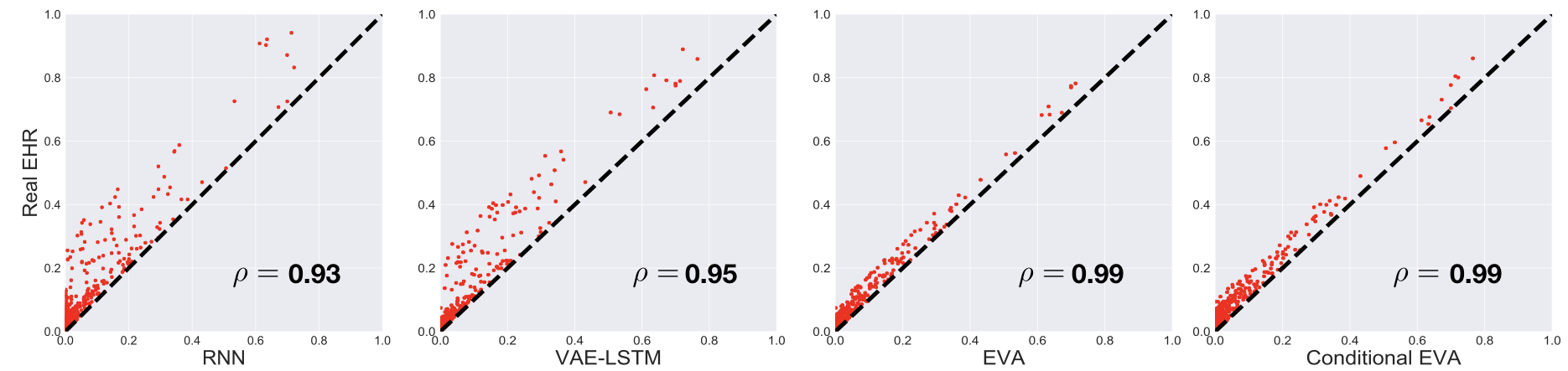}

\caption{\small{Additional scatter plots comparing the marginal (uni-gram) statistics of codes in real and synthesized EHRs.}}
\label{fig:unigram}
\end{figure*}

We have compared the marginal distribution of codes along with with the marginal statistics to evaluate how well \mname mimics real EHR data distribution\ref{fig:unigram}.

\subsection{Additional details for EVA$_c$ experiments}
To evaluate \cmname we generated several condition specific datasets. The conditions were --- Heart Failure, Acute Kidney Failure, Essential Hypertension, and Daibetes Mellitus. In order to determine which codes in the generated cohorts were relevant to the particular condition we used the following ICD9 mapping based on clinical input.
\begin{itemize}
\item Daibetes Mellitus: 250.00, 362.00, 357.00, 648.00, 249.00, 584.00
\item Heart Failure: 428.00, 402.00, 398.00, 404.00
\item Essential Hypertension: 401.00, 642.00
\item Acute Kidney Failure: 584.00, 669.00
\end{itemize}

\subsection{Informative latent space}
Informative latent space: We find that the using a feed forward architecture for the autoregressive likelihoods rather than a recurrent one helps alleviate the issue of KL collapse. For EVA, KL accounts for $22\%$ of the ELBO while for EVAc it accounts for $28\%$ of the ELBO. Additionally, Figure 3 in the supplement qualitatively demonstrates that the local latent variables (z) are informative and cluster based on patient condition.
\end{document}